\documentclass{article}
\PassOptionsToPackage{square,authoryear,comma}{natbib}
\usepackage[sglblindworkshop,final]{neurips_2026}
\workshoptitle{PALM: Personalized, Aligned, Long-Term Memory for AI Systems}

\usepackage[utf8]{inputenc}
\usepackage[T1]{fontenc}
\usepackage{amsmath,amssymb}
\usepackage{booktabs}
\usepackage{array}
\usepackage{graphicx}
\usepackage{xcolor}
\usepackage{tikz}
\usetikzlibrary{arrows.meta,positioning}
\usepackage{microtype}
\ifdefined\pdfinterwordspaceon\pdfinterwordspaceon\fi
\ifdefined\pdfglyphtounicode
\input{glyphtounicode}
\fi
\usepackage{url}
\usepackage[hidelinks]{hyperref}

\newcommand{\xattn}[1]{XAttn-#1}
\newcommand{\Wout}{W_{\mathrm{out}}}

\title{Not Your Diary, Just Your Topic:\\Grading the Wrong-User Control}

\author{%
  \normalfont Karina Romanova$^{1}$ \quad Igor Kiselev$^{2}$ \quad Vladimir Platonov$^{3}$ \\
  Viliana Devbunova$^{1}$ \quad Gregory Leleytner$^{4}$ \quad Adeliya Leleytner$^{5}$ \\
  Lina Veltman$^{6}$ \quad Aleksandr Nikolich$^{7}$ \\
  \normalfont $^{1}$Yandex \quad $^{2}$Accenture \quad $^{3}$RawModels \quad $^{4}$Independent Researcher \\
  \normalfont $^{5}$Institute for Information Transmission Problems \\
  \normalfont $^{6}$TBC Uzbekistan \quad $^{7}$Keenable AI%
}

\begin{document}
\maketitle

\begin{abstract}
A common test of personalization replaces a user's history with someone else's and measures the drop in accuracy. But this swap also changes the topics in the history, so the drop may reflect topic mismatch. We examine this problem with a frozen language model on LaMP-2 news categorization, using histories from similar users and synthetic histories assembled from other users' articles. The user's own history improves accuracy by 2.3 percentage points over a model trained without history. Removing history only at evaluation makes this gain appear larger, at 4.8 points. Category labels alone reproduce 86\% of the accuracy gap between the user's own history and a random user's history. With synthetic histories matched on labels and text similarity, the gap falls to 0.9 points. Across two user splits and three training seeds, the advantage over a similar user's history is 3.2 points for users seen during training and 0.65 points for unseen users, where the evidence for an advantage remains inconclusive. A small movie-tagging experiment with one training seed also finds no advantage over other users' histories. These findings show why personalization should be evaluated on unseen users, with histories of similar content and a model trained without history as comparisons. We also find that adding gated cross-attention improves on a prefix-only model under our training setup; whether this holds more broadly remains open.
\end{abstract}

\section{Introduction}\label{sec:intro}

Personalization systems are increasingly built the same way: freeze a capable decoder, compress what is known about a user into a small set of vectors, and inject those vectors somewhere in the forward pass \citep{ning2024userllm,liu2025personaplug,hebert2024persoma,huber2025e2p}. The compressed history is the memory. Whether the resulting system scores well and whether the decoder actually reads that memory are different questions, and benchmarks answer only the first.

The usual way to approach the second question is a \emph{wrong-user control}: replace the user's history with someone else's and measure the drop. LaMP reports gains even from randomly selected documents within the same user's profile \citep{salemi2024lamp}, but those documents still carry the user's topics. The random-history baselines of LaMP and OPPU \citep{tan2024oppu} draw instead from the wider profile corpus, which gives a non-personalized comparison. Read as a test of user-specific conditioning, however, a random replacement answers a weaker question than it appears to: another user's history differs in subject matter as well as in preferences, so the swap does not isolate identity. \citet{wu2024profile} come closest to the control this calls for, retrieving query-relevant items from other users' profiles on LaMP, where such semantic-only context helps on two tasks and hurts on four; PRIME \citep{zhang2025prime} replaces both profile text and user-specific LoRA weights with those of users of graded similarity and reports that every metric falls. Wu et al. explicitly distinguish semantic from personalization information; our question is how whole-profile swap contrasts change under stronger matching and user-disjoint evaluation.

\begin{figure}[t]
  \centering
  \includegraphics[width=\textwidth]{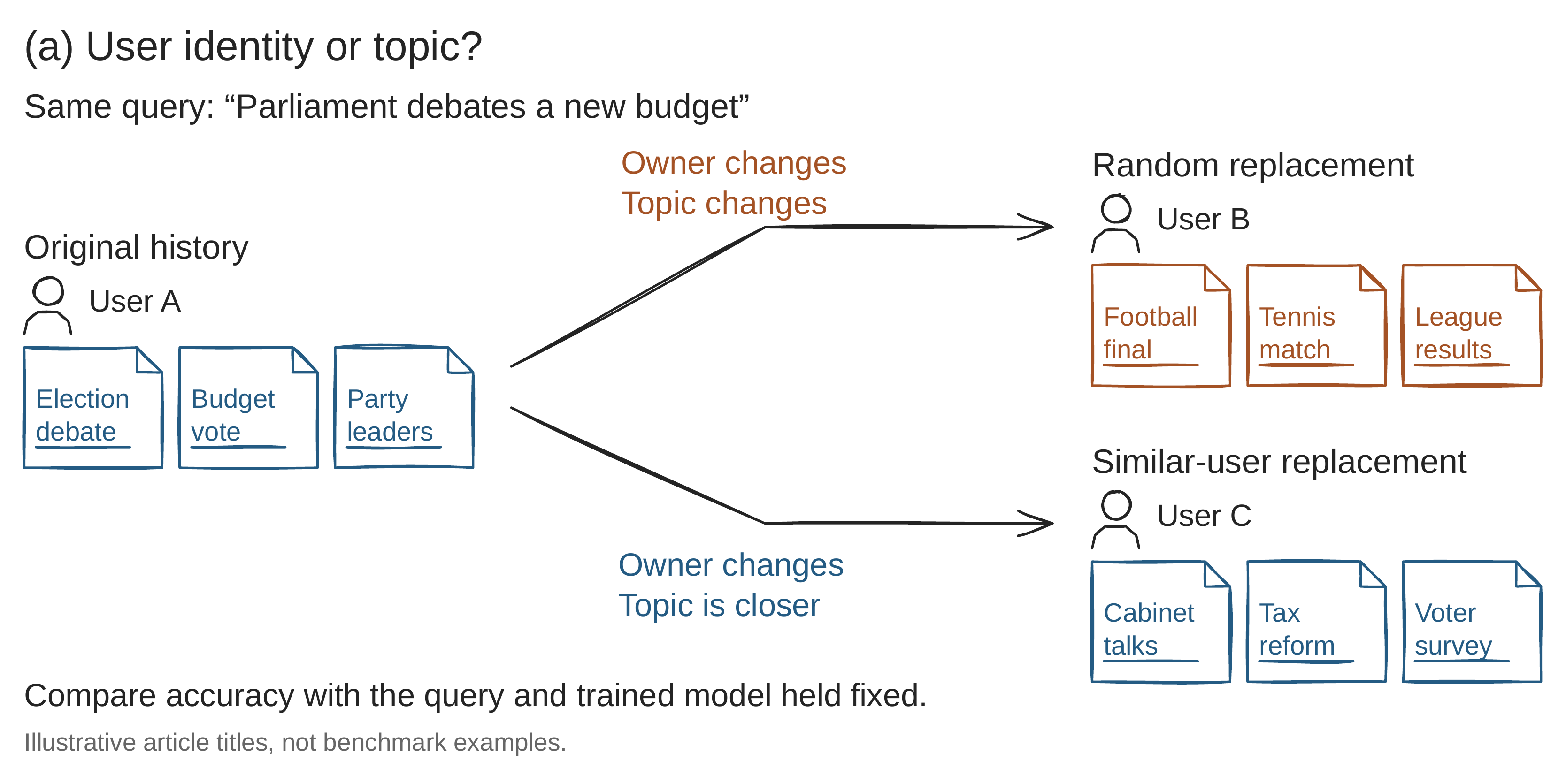}\\[6pt]
  \includegraphics[width=0.93\textwidth]{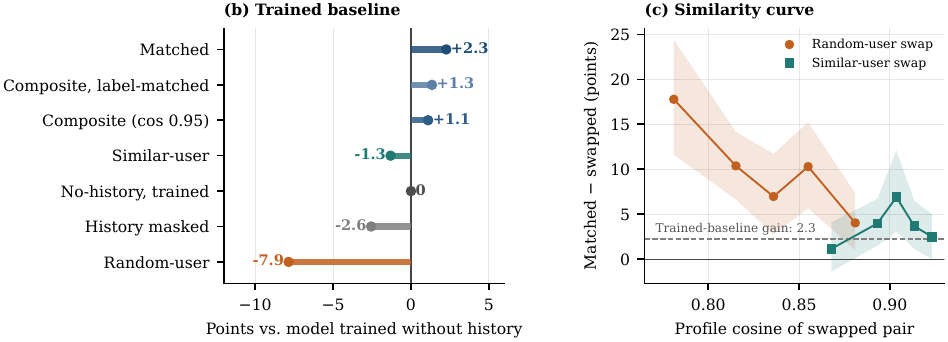}
  \caption{LaMP-2 news controls. (a) Illustrative histories show why a random-user swap mixes a change of owner with a change of topic. A similar-user history reduces content mismatch; the query and trained checkpoint stay fixed. (b) Three-seed mean accuracy relative to a separately trained no-history model. (c) Matched-minus-swapped accuracy against profile cosine, separately for random and similar users; bands resample users, retaining their repeated measurements. The dashed line is the 2.3-point gain over the trained comparator. The panels do not decompose a causal identity effect.}
  \label{fig:headline}
\end{figure}

We therefore study the measurement itself. The organizing idea comes from probing, where control tasks distinguish information recoverable from a representation from a probe's capacity to memorize \citep{hewitt2019control}. We \emph{grade} the wrong-user control: in place of one random replacement we use replacements of increasing similarity to the matched profile, first a similarity-matched real user and then synthetic profiles assembled from other users' items, and ask how much of the measured advantage survives each. The test bed is a family of adapters around a frozen decoder, each compressing a user's history into eight vectors. Two injection routes, a soft prefix and gated cross-attention, are compared in the main text and a third in Appendix~\ref{app:kv}. We also train a no-history comparator and evaluate on users excluded from training. These checks distinguish sensitivity to a history from a gain over a model that has learned to work without it (Figure~\ref{fig:headline}).

We make three empirical contributions:
\begin{itemize}
  \setlength{\itemsep}{2pt}
  \setlength{\parskip}{0pt}
\item We combine graded wrong-user replacements with a trained no-history reference and user-disjoint evaluation (\S\ref{sec:control}). Category controls explain much of the swap sensitivity, while the residual over strong negatives is small and uncertain on unseen users.
\item We test per-user concentration against nulls at the observed item counts and correct the scale comparison with paired permutation tests (\S\ref{sec:peruser}).
\item We compare a prefix with a prefix plus gated cross-attention, auditing output parsing and reporting non-shared parameter counts (\S\ref{sec:arch}). Deeper prefix slot encoders fit worse under this recipe; the comparison does not establish a capacity limit for prefixes.
\end{itemize}

\section{Setup and controls}\label{sec:setup}

\subsection{Model and injection routes}

A frozen BGE encoder maps each history item to an embedding. A separate, trainable BGE encoder represents the current input, and an eight-query, two-layer Q-Former combines the two into $Z \in \mathbb{R}^{8\times 768}$. The decoder and the history-item encoder stay frozen; the input encoder, the Q-Former and the injection modules train (Figure~\ref{fig:routes} in Appendix~\ref{app:kv} draws the three routes).

\textbf{Prefix.} $Z$ is projected to the decoder width and the eight vectors are prepended to the input embeddings, the route of \citet{liu2025personaplug,hebert2024persoma,huber2025e2p}.

\textbf{\xattn{$k$}.} In addition to the prefix, $k$ selected decoder layers receive a residual update
\begin{equation}
  h' = h + \tanh(\alpha)\,\Wout\,\mathrm{CrossAttn}\big(\mathrm{LN}(h), Z\big),
  \label{eq:xattn}
\end{equation}
with $\alpha_0 = 1.5$ and $\Wout = 0$ at initialization unless stated otherwise. The main cross-attention comparisons use this hybrid: it retains the input prefix and adds gated branches, so the comparison between the two routes is between a prefix alone and a prefix plus cross-attention. Which layers receive the update is swept in \S\ref{sec:sweep}.

\subsection{History conditions}\label{sec:conditions}

We call a replacement history a \emph{negative} and the matched-minus-negative accuracy difference a \emph{contrast}. The core battery has five history conditions, supplemented by category controls and a trained comparator. \textbf{Matched} uses the profile attached to the example. \textbf{Random-user} uses the history of another user chosen by a deterministic derangement, so that no user maps to itself; the mapping is fixed across training, validation and model seeds, and this is the standard wrong-user control. \textbf{Similar-user} uses the other user assigned by the derangement that maximizes total profile cosine subject to having no fixed point (\S\ref{sec:battery}). \textbf{Composite} is a synthetic profile of six items drawn from other users so that its centroid matches the target profile's (\S\ref{sec:composite}). \textbf{History masked} (called No-history in the earlier experiments) masks every history item at evaluation while the Q-Former still attends to the current input; on the fixed checkpoints of \S\ref{sec:control}--\ref{sec:peruser} it is a counterfactual masking condition on the matched-trained model. We additionally train a \textbf{No-history, trained} model with this mask throughout training and evaluation, following the input-ablation logic of \citet{kaushik2018reading}.

Random-user and History masked answer different questions: the first preserves the marginal distribution of histories but breaks their relation to the current query, and measures sensitivity to a mismatched history; the second removes the history and measures whether the supplied one helps relative to none. Neither is guaranteed to lie within the training distribution, and their difference is itself informative.

The controls can be applied counterfactually, fixing a trained checkpoint and changing only its history, or by training a model under each condition; the main battery fixes the matched-trained checkpoint, whereas the trained no-history, similar-history training, and user-fold experiments explicitly retrain models.

\subsection{Data, protocol and conventions}\label{sec:protocol}

\textbf{Task.} The primary case study is the deprecated news-categorization variant of LaMP-2 \citep{salemi2024lamp}: 15 classes, 1,605 validation examples, 323 validation users, and the first six items of each available profile. All 323 profiles recur in the full 8,090-row training file; the first 8,000 rows used by the main recipe cover 314 of them. This is a time-based split, not a test on new users. Article categories are available in the raw data but were omitted from the original serialized histories; we recover them for the label controls. The benchmark has since replaced this task with movie tagging,\footnote{\url{https://lamp-benchmark.github.io/download}} and ``LaMP-2'' below refers to the news task only. Always predicting the majority class scores 0.476, a reference for the task. History items are serialized with a constant \texttt{MOVIE:} label inherited from the other LaMP-2 variant; removing it has little effect in the seed-42 serialization check.\footnote{Retraining \xattn{9} without the label gives 0.7888 against 0.7882 at seed 42, an order of magnitude below the seed standard deviation.}

\textbf{Decoders.} The frozen decoder is Qwen2.5-3B-Instruct for the control battery and the injection-site comparison, Qwen2.5-0.5B-Instruct for a second scale point, and Qwen2.5-1.5B for the initialization ablation. Both routes train at the learning rate a six-point sweep selects for each, $3\times10^{-5}$ (Appendix~\ref{app:hyper}), which lists the training settings.

\textbf{Output validity.} A prediction counts as valid only when the full stripped generation exactly matches a task label. We report output validity alongside accuracy, because on a format-sensitive task a swap control otherwise registers format priming as user specificity. The main control and architecture comparisons use this audited harness, including retrained prefix rows; the earlier initialization ablation is retained as a gradient-failure diagnostic. An unaudited parser depresses the soft prefix by 4.5 accuracy points and 28 macro-F1 points while reproducing cross-attention exactly (Appendix~\ref{app:unaudited}). The harness, the analysis scripts and the saved per-item predictions are in the supplement (Appendix~\ref{app:repro}).

\textbf{Uncertainty.} The main paired contrasts use a two-level bootstrap over model seeds and users, motivated by the system/input resampling of \citet{deutsch2021statistical}; our nested seed/user procedure differs from their crossed resampling. With three seeds there are only ten distinct seed multisets; Appendix~\ref{app:perseed} therefore reports $t$-intervals with two degrees of freedom alongside the bootstrap. We also resample constructions for the composite and seed--fold cells followed by users for the user split. These exploratory intervals are uncorrected for multiple comparisons. Repeating the 3B prefix at a fixed seed moved accuracy from 0.7445 to 0.7346 under non-deterministic fp16 training, so small differences between separately trained models warrant caution. Accuracies are fractions; contrasts are in accuracy points unless stated otherwise.

\section{Whose history, or merely which topic?}\label{sec:control}

A random derangement replaces a user's history with somebody else's, and the replacement is usually about different things. Any penalty it produces therefore mixes two effects, that the history is not this user's and that it is off-topic, and random-shuffle controls cannot separate them. We add harder negatives that reduce the mismatch in embedding space.

\subsection{Graded negatives on LaMP-2}\label{sec:battery}

The battery runs on three independently trained \xattn{9} checkpoints on Qwen2.5-3B (seeds 7, 42 and 123; the strongest configuration of \S\ref{sec:arch}), each evaluated under every condition with the model fixed and only the history changing. The Similar-user derangement raises the mean cosine of a swapped pair from 0.834 to 0.900, with the least similar assigned pair still at 0.806. Composites are built greedily from the 400 items nearest the target (Appendix~\ref{app:repro}) and reach a cosine of 0.948; a second targets cosine 0.90 at each greedy step over all other-user items to compare with the Similar-user condition at equal similarity.

\begin{table}[t]
\centering\footnotesize
\caption{LaMP-2 news accuracy. All history interventions use the same three matched-trained checkpoints; the last row trains and evaluates with history masked. Labels-only inputs contain the six categories without article text. Valid is the mean exact-label output rate. Intervals for paired contrasts are in Appendix~\ref{app:perseed}.}
\label{tab:battery}
\begin{tabular}{@{}lrrrrr@{}}
\toprule
Condition & Seed 7 & Seed 42 & Seed 123 & Mean & Valid \\
\midrule
Matched & 0.7975 & 0.7882 & 0.7950 & 0.7936 & 0.997 \\
Composite, label-matched & 0.7844 & 0.7844 & 0.7844 & 0.7844 & 0.997 \\
Composite, $\cos=0.95$ & 0.7857 & 0.7801 & 0.7801 & 0.7819 & 0.998 \\
Composite, $\cos=0.90$ & 0.7595 & 0.7595 & 0.7558 & 0.7583 & 0.997 \\
Similar-user & 0.7551 & 0.7645 & 0.7539 & 0.7578 & 0.996 \\
Labels only, own & 0.7682 & 0.7688 & 0.7626 & 0.7666 & 0.997 \\
Labels only, random user & 0.6897 & 0.6891 & 0.6598 & 0.6795 & 0.998 \\
History masked & 0.7427 & 0.7439 & 0.7495 & 0.7454 & 0.996 \\
Random-user & 0.6903 & 0.6997 & 0.6872 & 0.6924 & 0.997 \\
\midrule
No-history, trained & 0.7769 & 0.7670 & 0.7688 & 0.7709 & 0.999 \\
\bottomrule
\end{tabular}
\end{table}

The trained comparator changes the interpretation of the battery (Table~\ref{tab:battery}). Let A be training and evaluation with matched histories, B the same checkpoint evaluated with its history masked, and C training and evaluation with the mask. A exceeds B by 4.8 points but C by only 2.3 (2.06, 2.12 and 2.62 across seeds). C exceeds B by 2.6 points. Thus A$-$B measures dependence on history at a fixed checkpoint and overstates the gain relative to a model trained to operate without it. Random-user is 7.9 points below C, accounting for 78\% of the 10.1-point A-minus-random contrast; relative to B that fraction is 52\%. Similar-user is 1.3 points below C, while the closest composite is 1.1 above it. These baseline contrasts do not identify a causal share of user identity.

\subsection{The contrast as a function of similarity}\label{sec:curve}

The two derangements provide 646 distinct user--donor pairs evaluated on three checkpoints. The six pooled equal-count bins give contrasts of 14.3, 9.6, 8.6, 0.1, 3.7 and 2.7 points at mean cosines 0.793, 0.834, 0.863, 0.887, 0.905 and 0.920. Figure~\ref{fig:headline}c instead separates the swap types and clusters uncertainty by user. For Random-user, the fitted slope is $-1.2$ points per 0.01 cosine, with 95\% interval $[-2.0,-0.4]$; for Similar-user it is $-0.2$ $[-0.9,0.8]$. Similarity is associated with a smaller penalty among random swaps, but the similar-user range shows no clear slope. This cross-sectional curve neither establishes a within-user effect nor supports extrapolation to a perfect match.

\subsection{Building the negative}\label{sec:composite}

The closest composite reaches mean cosine 0.948 and leaves a 1.2-point matched advantage, with bootstrap interval $[0.4,1.9]$. Rebuilding under E5 instead of BGE leaves 1.7 points; the Similar-user contrast becomes 4.0 rather than 3.6. A complete grid of three seeds and three greedy constructions gives a mean residual of 1.4 points, with a three-level bootstrap interval of $[0.7,2.1]$. The standard deviations across construction means and seed means are 0.21 and 0.46 points.

We also constrain each donor item to match the original item's category at the same position. This joint composite reaches cosine 0.941 with 100\% category agreement, no matching failures, and 5.8 donors per profile. Its residual is 0.9 points: the bootstrap interval is $[0.1,1.8]$ and the three-seed $t$-interval is $[-0.3,2.1]$. Joint matching changes both the label balance and the achievable cosine, so the reduction from 1.2 to 0.9 does not isolate a label effect. Neither construction separates identity from coherence or shared preferences.

\subsection{What the residual tracks}\label{sec:labels}

The matched-minus-masked gain is concentrated in the 850 queries whose category occurs in the six-item profile: 9.7 points, against $-0.7$ on the other 755. The corresponding contrasts with Similar-user are 4.8 and 2.2 points, and with the closest composite 3.2 and $-1.1$. For Similar-user the gap is 9.6 points when the category occurs only in the matched profile ($n=236$), 3.0 when it occurs in both ($n=614$), and $-4.6$ when it occurs only in the donor ($n=183$). These strata use the reference category for analysis, not for constructing the controls.

Six category labels without article text score 0.7666 for the user's own profile and 0.6795 for a random donor. Their 8.7-point contrast reproduces 86\% of the full random-swap contrast, although labels-only inputs were not used in training. Own labels exceed the masked condition by 2.1 points but remain 0.4 below the trained no-history model. A majority vote over the six profile labels scores only 0.436, against the global majority's 0.476; a lookup of each user's most frequent training answer scores 0.675. The results implicate category information without showing that a simple profile-majority rule explains the model.

We replace nested subsets of $k=0,\ldots,6$ history positions with items from a fixed donor (Appendix~\ref{app:algorithms}). Replacing one or two items changes accuracy by at most 0.52 points for all three donor types. Full replacement costs 10.1 points for a random user, 3.6 for a similar user, and 0.9 for the label-matched composite (Figure~\ref{fig:dose}). The increasing separation between donor types shows sensitivity to replacement content; it is not an additive decomposition of topic and identity.

\subsection{Seen and unseen users}\label{sec:unseen}

We split the 323 users into folds of 161 and 162, each contributing 4,045 training rows, and train on each fold for 6,000 steps at each of three seeds. Every checkpoint is evaluated on both its training users and the held-out users under the full battery. Matched accuracy averages 0.784 for seen users and 0.738 for unseen users. Table~\ref{tab:unseen} reports all six cells, with intervals resampling cells and then users within each group.

\begin{table}[t]
\centering\small
\caption{Matched minus each control, accuracy points (95\% bootstrap intervals), three seeds and two user folds. The masked condition uses the same checkpoint; no separate no-history model was trained within these folds.}
\label{tab:unseen}
\begin{tabular}{@{}lrr@{}}
\toprule
Control & Seen users & Unseen users \\
\midrule
History masked & $3.7\ [1.7,5.6]$ & $0.8\ [-0.7,2.3]$ \\
Random-user & $11.0\ [8.0,14.1]$ & $7.9\ [5.4,10.4]$ \\
Similar-user & $3.2\ [0.9,5.7]$ & $0.65\ [-1.1,2.2]$ \\
Composite & $0.6\ [-0.6,2.0]$ & $0.4\ [-0.6,1.5]$ \\
Composite, label-matched & $-0.1\ [-1.2,1.0]$ & $-0.4\ [-1.5,0.5]$ \\
\bottomrule
\end{tabular}
\end{table}

The similar-user residual is smaller on unseen users, and all their strong-negative intervals include zero. Random swapping still produces a large penalty. These findings are consistent with profiles supplying a learned user prior in the time split, but do not establish that all residual signal is memorization: the fold models see fewer training users, and the intervals allow small benefits on unseen users. Training on a fixed Similar-user assignment gives accuracy 0.789, close to matched training's 0.794; evaluating those models with the users' own histories reduces accuracy to 0.761. Because the assignment is a bijection, a donor profile can serve as a stable user key. This experiment cannot distinguish such recognition from learning the donor's content distribution.

On the active LaMP-2 movie-tagging task, a seed-42 pilot evaluated on 692 queries from unseen users scores 0.5000 with matched history, 0.5014 with Similar-user, 0.5029 with a composite, 0.4957 with Random-user and 0.4509 with masking. Mean donor cosines are 0.977 and 0.994 for Similar-user and composite. The matched history has no observed advantage over the other histories, despite its 4.9-point advantage over masking. One seed and no trained no-history comparator limit this pilot.

We attempted the replacement battery on the no-history-trained model as a negative control. Its seed-7 checkpoint generated concatenated labels under matched history, with zero valid outputs. We stopped that battery: unmasking a branch never trained on history does not provide a usable no-user-mechanism control.

\subsection{Not an artifact of compression, scale or schedule}\label{sec:robust}

The ordering is not confined to the compressed memory. The same four histories supplied as text to the frozen decoder, with no training and no bottleneck, reproduce it (Appendix~\ref{app:text}), with a Matched-minus-Similar-user contrast of 2.3 points against the adapter's 3.6; so does a frozen Qwen2.5-0.5B trained to the same recipe, with Matched minus Similar-user at 4.5 points $[2.6, 6.6]$, the interval for a similar user's advantage over masked history again spanning zero, and a random one 7.6 points below it (Appendix~\ref{app:text}). One contrast that looks like a scale effect is not: a 0.5B checkpoint trained on a shorter schedule at a higher learning rate shows almost no penalty under any replacement, and matching the recipe removes the difference (Appendix~\ref{app:hyper}).

An article-text audit finds one exact query article in its own profile among 1,605 queries. Removing the 9.7\% nearest to their profiles by article-text similarity changes the Similar-user contrast from 3.6 to 3.7 points and the composite contrast from 1.2 to 0.8 $[0.1,1.6]$. Near-duplicates do not explain the main Similar-user result.

\section{The per-user decomposition does not survive its own null}\label{sec:peruser}

An average says nothing about who receives it, and thresholded summaries invite the objection that the threshold does the work \citep{schaeffer2023mirage}, so we report continuous statistics. Decomposing the Matched-minus-Random-user effect over the 323 validation users gives a Gini coefficient of the per-user effect magnitude of $0.727 \pm 0.016$ at 0.5B and $0.761 \pm 0.017$ at 3B, with the top decile of users carrying 46\% and 51\% of the total effect mass. Such figures are routinely read as an effect carried by a small minority of users; the reading does not survive a null.

$\mathrm{Gini}(|\Delta|)$ runs high by construction whenever $\Delta$ is noisy around a mean near zero, and that is the regime here: each user contributes about five validation items, so $\Delta_u$ is a difference of two means over five binary outcomes. Two permutation nulls on the same users and item counts test the reading: the \emph{zero-effect} null exchanges the two condition labels within each item, removing any systematic effect while preserving how often the conditions disagree, and the \emph{homogeneous-effect} null redraws each user's items from the pooled item-level outcome pairs, keeping the population effect and destroying only user identity (Table~\ref{tab:null}).

\begin{table}[t]
\centering\small
\caption{Per-user decomposition against permutation nulls at matched item counts: Gini of the per-user effect magnitude and the standard deviation of the per-user effect across users (not reported under the zero-effect null). The observed Gini sits next to the homogeneous-effect null and below the zero-effect null; only dispersion exceeds its null.}
\label{tab:null}
\begin{tabular}{@{}llccc@{}}
\toprule
Decoder & Statistic & Observed & Homogeneous-effect null & Zero-effect null \\
\midrule
0.5B & Gini$(|\Delta_u|)$ & $0.727 \pm 0.016$ & 0.720 & 0.760 \\
0.5B & SD$(\Delta_u)$ & 0.352 & 0.318 & -- \\
3B & Gini$(|\Delta_u|)$ & $0.761 \pm 0.017$ & 0.758 & 0.791 \\
3B & SD$(\Delta_u)$ & 0.323 & 0.294 & -- \\
\bottomrule
\end{tabular}
\end{table}

The observed Gini sits close to the homogeneous-effect null and below the zero-effect null at both scales, with per-seed permutation probabilities for the Gini of 0.17 to 0.52 at 0.5B and 0.07 to 0.80 at 3B, so these values do not establish concentrated personalization. Dispersion does leave a residue: the observed standard deviation across users exceeds the homogeneous null in all six runs, with three of the six per-seed probabilities below 0.05 before any multiplicity correction. This is exploratory evidence of heterogeneity, not an account of which users benefit; that needs more items per user, and papers that decompose personalization gains over users should report items per user alongside the statistic.

The scale comparison requires a paired test because effects are discrete and the same users recur across seeds. Swapping the two decoder-scale labels within each user gives Kolmogorov--Smirnov permutation probabilities of 0.45, 0.010 and 0.0035, and Cram\'er--von Mises probabilities of 0.74, 0.0035 and 0.0010. Summing the seed-level statistics gives $p=0.0015$ for each test under independent within-user swaps in each seed. The distributions are distinguishable under this permutation scheme, contrary to the asymptotic KS reading. Their mean Matched-minus-Random-user contrasts are 12.8 points at 0.5B and 10.1 at 3B under the same recipe. The earlier, larger scale difference was affected by the shorter training schedule; that does not remove this remaining difference.

At five evaluation items per user the per-user estimate is mostly noise whatever the model does; aggregates can hide structure when the cells are estimable \citep{deng2026substrate}, but here the decomposition was not identifiable from the data we had.

\section{Where the audit finds nothing to grade}\label{sec:othertasks}

LaMP-2 is the one task of five on which the battery finds anything to decompose. On the other four the measured contrasts are small (Table~\ref{tab:othertasks}): the differences are the size of the seed spread, and on LaMP-3 the Random-user-minus-History masked difference agrees in sign on one seed of three. One degenerate LaMP-3 run (the majority rating on 1,596 of 1,605 items) is a training failure, and excluding it changes no conclusion. Output validity is flat across conditions (0.999 in range on LaMP-3), so format priming is not at work. With nothing separating Matched from Random-user there is nothing to grade by similarity, so we built no topic-matched negatives for these tasks.

\begin{table}[t]
\centering\footnotesize
\caption{The four other LaMP tasks under the three conditions. Three seeds for LaMP-3/5 and two for LaMP-1/4; history controls are counterfactual evaluations of each matched-trained checkpoint. Lower is better for mean absolute error (MAE); higher is better otherwise. One metric per task is reported; the benchmark also defines RMSE for LaMP-3 and ROUGE-L for LaMP-4 and LaMP-5. Differences between conditions are of the same size as the seed spread.}
\label{tab:othertasks}
\begin{tabular}{@{}llcccc@{}}
\toprule
Task & Metric & Matched & History masked & Random-user & Matched $-$ Random-user \\
\midrule
LaMP-1, citation choice & Accuracy & 0.6707 & 0.6757 & 0.6700 & $+0.0007$ \\
LaMP-3, product rating & MAE $\downarrow$ & 0.436 & 0.447 & 0.446 & $-0.010$ \\
LaMP-4, news headline & ROUGE-1 & 0.1923 & 0.1886 & 0.1859 & $+0.0064$ \\
LaMP-5, scholarly title & ROUGE-1 & 0.5008 & 0.4966 & 0.5006 & $+0.0002$ \\
\bottomrule
\end{tabular}
\end{table}

Supplying the histories as text to the frozen decoder does not recover a gain either: both tested tasks get worse, mostly through output format (Appendix~\ref{app:text}), which leaves the task, the metric and the training recipe as candidate explanations and neither establishes that the tasks lack personalization signal nor rules out a compression bottleneck.

\section{Accuracy against trainable parameters across injection routes}\label{sec:arch}

\begin{figure}[t]
\centering
\includegraphics[width=0.60\textwidth]{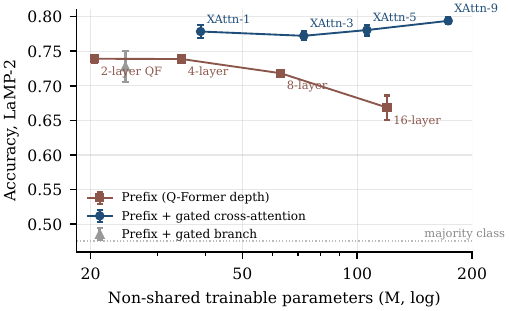}
\caption{Accuracy against non-shared trainable parameters, excluding the shared 109M-parameter input encoder. Error bars are standard deviations over three seeds. Cross-attention models retain the prefix; these are not exactly budget-matched route comparisons. Full metrics appear in Table~\ref{tab:arch}.}
\label{fig:capacity}
\end{figure}

The battery above takes the injection route as given; this section asks what adding a gated cross-attention read path to a prefix model buys per trainable parameter. Since every cross-attention model keeps the prefix (\S\ref{sec:setup}), this is prefix against prefix plus cross-attention, not one injection site against another; \citet{ning2024userllm}, for example, compare a length-10 soft prompt with cross-attention under full fine-tuning, without reporting paired trainable parameter counts for those arms.

\subsection{Accuracy and trainable-parameter comparisons}\label{sec:sweep}

We sweep the number of injected layers on the audited harness with three seeds and all 1,605 validation items (Table~\ref{tab:arch}, lower block). The first injected layer is worth 3.9 points of accuracy and 7.2 of macro-F1 for 14\% more trainable parameters; the remaining eight layers add a further 1.5 and 4.7 for 90\% more. The gain from one to nine layers is about 40\% of the first step; three seeds and the repeat-run variability of \S\ref{sec:protocol} do not justify calling the curve flat. \xattn{1} beats the prefix at $1.14\times$ its total trainable parameters and about $1.9\times$ its non-shared parameters once the shared 109M-parameter input encoder is set aside (38.9M against 20.5M); it also beats the 4-layer prefix (34.7M) by 4.0 points and the 8- and 16-layer prefixes (63.0M, 119.7M) by more, and nine added layers reach 173.2M. These comparisons trace a trade-off; they do not isolate a route effect at exactly matched capacity. Validity holds at 0.995--0.997 across the injected rows, 1.1 points above the prefix. Tuning the learning rate for both routes widens the gap rather than closing it (Appendix~\ref{app:hyper}).

\citet{jeong2026memory} compare six memory methods on a frozen GPT-2 under shared memory-size settings, with different trainable parameter counts. At the smaller setting, cross-attention, Hebbian memory and sparse slot writing outperform the other methods; the gap narrows at the larger setting. Both reading and writing mechanisms vary, so this comparison does not isolate the injection route. A graph memory read by cross-attention also beats a prefix on knowledge-graph completion \citep{liu2025graph}.

Deepening the prefix's Q-Former from two to sixteen layers lowers accuracy from 0.7389 to 0.6685 while median training loss rises. Increasing prefix length from 8 to 64 slots at seed 42 does not close the gap. These are failures to fit under the tested recipe, with optimization and expressivity still unresolved (Appendix~\ref{app:capacity}). A separate initialization ablation confirms that setting both a residual gate and its output projection to zero blocks branch gradients; a parallel prefix can conceal that failure (Appendix~\ref{app:init}).

\section{Related work}\label{sec:related}

User-LLM injects user representations through cross-attention \citep{ning2024userllm}. Persona-Plug, PERSOMA, Embedding-to-Prefix and REPREC map a user representation to learned soft tokens in front of a frozen decoder \citep{liu2025personaplug,hebert2024persoma,huber2025e2p,kavuru2026reprec}, the route we call the prefix (input-level prompt tuning; \citealp{lester2021power}). Flamingo and LLaMA-Adapter leave one gradient path open at initialization \citep{alayrac2022flamingo,zhang2024llamaadapter}; the deadlock of Appendix~\ref{app:init} needs both closed \citep{khemais2026surgery}.

The control design descends from trained input-ablated baselines \citep{kaushik2018reading}, control tasks \citep{hewitt2019control} and explicit positive and negative controls \citep{mu2023gist}. That an off-target context can cost more than no context has been observed for retrieved text \citep{petroni2020context,yoran2024robust}. Matched or graded replacements have precedents (Appendix~\ref{app:position} sets them side by side): \citet{wu2024profile} retrieve query-relevant items from other users' profiles on LaMP as a semantic-only control, PRIME \citep{zhang2025prime} swaps in the memories of users of graded similarity, and, concurrently with this work, ENOUGH \citep{liu2026enough} scores each record against content-matched other-user replacements as a training signal; \citet{zheng2026weighting} likewise find a shuffle control necessary but not sufficient in multimodal recommenders. We combine whole-profile controls with category matching, a trained no-history reference, and evaluation on users excluded from training. The within-user replacement curve complements the cross-sectional similarity curve; neither is a causal decomposition of personalization. Audits that decompose a single personalization score are adjacent \citep{deng2026substrate}, as is the shortcut-and-confound protocol of \citet{chen2026elicitation} for memory-policy classification.

\section{Limitations}\label{sec:limits}

The main case study uses a deprecated news task, six-item truncations of longer profiles, and a time split with extensive user overlap. Category matching and the near-duplicate audit reduce specific confounds but do not hold topic, style, coherence and preferences separately fixed. User-disjoint training changes the available training population, and its intervals include modest benefits from matched history. The seed--fold bootstrap treats cells as resampling units even though users recur across cells; its intervals should be read as exploratory, not as an equivalence test. The movie-tagging pilot has one seed and lacks a trained no-history comparator. The architecture comparisons are not exactly budget-matched, rely on one recipe, and leave the prefix's optimization failure unresolved. Longer profiles, other decoders, and deployment utility, latency and privacy remain outside this study.

\section{Conclusion}\label{sec:conclusion}

On LaMP-2 news, a 10.1-point random-swap penalty accompanies only a 2.3-point gain over a model trained without history. Category-only inputs reproduce much of the swap contrast, and matched histories show small, uncertain advantages over strong negatives on unseen users. Evaluating on users excluded from training and reporting a trained no-history reference are therefore essential checks before interpreting a wrong-user penalty as evidence of user-specific memory. Graded replacements, output-validity audits and per-user nulls then show which parts of that interpretation the data can support.

\clearpage
\bibliographystyle{plainnat}
\bibliography{refs}

\appendix

\section{Reproducibility details}\label{app:repro}

The supplement contains the training and evaluation harness, the analysis scripts, and saved run outputs with per-item predictions for the core comparisons; its README lists archive coverage. The Random-user mapping hashes the complete profile and applies a seeded derangement over users, shared between training and validation. Prediction files store the item identifier, current user, history user, raw generation and reference, which is what makes the decomposition of \S\ref{sec:peruser} possible after the fact. Result files also store source checksums, output-validity audits, training traces, peak GPU memory and the control-mapping checksum. The classification evaluator compares the full stripped generation with the labels found in the complete public train and validation splits, without substring matching.

Six details the audit turns on. The prompt is the benchmark's own input text, tokenized with special tokens and truncated to the decoder's window, preceded in embedding space by the neutral system string ``Answer concisely.''; training concatenates prompt, answer and end-of-sequence with the prompt positions masked out, and evaluation left-pads the prompt alone. The random derangement draws \texttt{torch.randperm} under the control seed and redraws up to a hundred times until no user maps to itself, falling back to a roll by one, which never fired. The similarity-matched derangement solves the assignment problem on the negated cosine matrix with \texttt{scipy.optimize.linear\_sum\_assignment}, setting the cosine diagonal to $-\infty$ before negation (cost $+\infty$). The composite negative selects items greedily from the four hundred items ranked nearest the target, at each step taking the candidate that brings the running centroid closest to it; repeated constructions replace that argmax with a uniform draw among the five best candidates at each step, seeded per user by the run seed plus the user's index, which is the stochastic element of the construction. The binned curve uses equal-count bins within each swap type, with user-clustered intervals; the earlier pooled six-bin summary is retained in \S\ref{sec:curve}. Every bootstrap resamples model seeds with replacement and then users within the drawn seed, weighting each user by its own item count.

The revision's 110 result files record 23.0 GPU-hours: 8.8 for training and 14.2 for evaluation, excluding four training attempts interrupted by out-of-memory errors. Runs used a shared V100-SXM2-32GB, PyTorch 2.5.1+cu118, Transformers 4.57.6, NumPy 1.26.4 and SciPy 1.17.1. Logged durations include contention on that device. The new models retain the 3B recipe unless specified: three seeds, 6,000 steps, learning rate $3\times10^{-5}$. User-fold models use all 4,045 training rows in their fold.

\section{Additional control results}\label{app:controls}
\begin{figure}[h]
\centering
\begin{minipage}{0.49\textwidth}
\includegraphics[width=\linewidth]{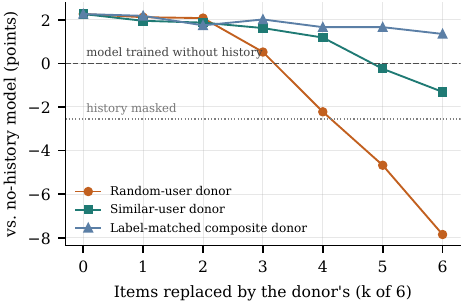}
\end{minipage}\hfill
\begin{minipage}{0.49\textwidth}
\includegraphics[width=\linewidth]{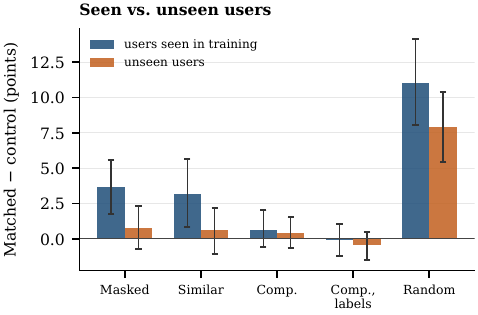}
\end{minipage}
\caption{Left: within-user replacement of $k$ out of six items, three-seed means relative to the trained no-history model; the dotted line is masking on the matched-trained model. Lines connect measured doses and carry no uncertainty band. Right: Matched-minus-control contrasts for seen and unseen users, with seed--fold/user bootstrap intervals (Table~\ref{tab:unseen}).}
\label{fig:dose}
\end{figure}

The dose curves use the same position order and donor for every $k$ within a user. The changes from the matched condition at $k=1,\ldots,6$ are $-0.15,-0.19,-1.74,-4.49,-6.94,-10.11$ points for Random-user; $-0.31,-0.39,-0.64,-1.08,-2.51,-3.57$ for Similar-user; and $-0.08,-0.52,-0.25,-0.60,-0.60,-0.91$ for the label-matched composite. Small doses have little effect, but the composite curve is not monotone.

\section{Architecture details}\label{app:capacity}
\begin{table}[h]
\centering\footnotesize
\setlength{\tabcolsep}{4.5pt}
\caption{Injection route against trainable parameters on LaMP-2, frozen Qwen2.5-3B, learning rate $3\times10^{-5}$, three seeds, sample standard deviations (for macro-F1 where recorded). Upper block: the prefix's Q-Former depth scaled at fixed conditioning length, and the prefix with a gated zero-initialized branch (\S\ref{sec:axes}). Lower block: \xattn{$k$} with injected layers \texttt{range(18, 36, s)} of the 36-layer decoder; strides $s = 18, 8, 4, 2$ give $k = 1, 3, 5, 9$. Each injected layer adds 16,789,505 trainable parameters; the first also adds a 1,579,008-parameter projection of $Z$. Train loss: median per-step loss over the final third of training (Appendix~\ref{app:loss}), not reported for \xattn{3} and \xattn{5}.}
\label{tab:arch}
\begin{tabular}{@{}lrcccccc@{}}
\toprule
Model & Trainable & $\times$prefix & Accuracy & Macro-F1 & Valid & Train loss \\
\midrule
Prefix (2-layer Q-Former) & 129,970,944 & 1.00 & $0.7389 \pm 0.0059$ & $0.4896 \pm 0.0044$ & 0.984 & 0.0041 \\
Prefix, 4-layer & 144,146,688 & 1.11 & $0.7385 \pm 0.0019$ & 0.484 & 0.981 & 0.0039 \\
Prefix, 8-layer & 172,498,176 & 1.33 & $0.7180 \pm 0.0013$ & 0.468 & 0.982 & 0.0126 \\
Prefix, 16-layer & 229,201,152 & 1.76 & $0.6685 \pm 0.0174$ & 0.375 & 0.973 & 0.0283 \\
Gated prefix, 2-layer & 134,165,249 & 1.03 & $0.7275 \pm 0.0221$ & 0.478 & 0.979 & 0.0087 \\
\midrule
\xattn{1} & 148,339,457 & 1.14 & $0.7782 \pm 0.0097$ & $0.5614 \pm 0.0114$ & 0.995 & 0.0011 \\
\xattn{3} & 181,918,467 & 1.40 & $0.7720 \pm 0.0065$ & $0.5651 \pm 0.0115$ & 0.997 & -- \\
\xattn{5} & 215,497,477 & 1.66 & $0.7803 \pm 0.0078$ & $0.5772 \pm 0.0182$ & 0.997 & -- \\
\xattn{9} & 282,655,497 & 2.17 & $0.7936 \pm 0.0048$ & $0.6084 \pm 0.0143$ & 0.997 & 0.0012 \\
\bottomrule
\end{tabular}
\end{table}

\subsection{Deepening the slot encoder makes the prefix worse}\label{sec:prefix}

Would a soft prefix given the same budget close the gap? Both routes share the 109M-parameter input encoder, so we scale the Q-Former depth at fixed input encoder and conditioning length (Table~\ref{tab:arch}, upper block).

Accuracy falls monotonically as the Q-Former deepens, from 0.7389 to 0.6685 across a $1.76\times$ increase in trainable parameters, 7.0 points against seed standard deviations of 0.002 to 0.017, and \xattn{1}, with fewer trainable parameters than the 8-layer prefix, outscores every prefix row. Loss of output format does not explain it: validity falls by 1.1 points over a range in which accuracy falls by seven.

Diminishing gains from longer prompts or larger prefix parameter budgets have been observed in specific settings \citep{li2021prefix,lester2021power,he2022unified}; \citet{li2021prefix} attribute it to overfitting, since longer prefixes lower their training loss. Our training loss moves the other way: its median rises sevenfold as capacity grows, and every injected model fits the training set better than every soft prefix. We conjecture that the deeper prefixes fail to fit rather than overfit; an expressivity ceiling \citep{petrov2024prompting} and plain optimization failure are both consistent with the gradient diagnostics of Appendix~\ref{app:grad}, the usual remedies for the latter (a reparameterization network, a separate prefix learning rate) were not tested, and the result therefore concerns this training recipe rather than soft prefixes as such.

\subsection{Two axes the parameter count cannot settle}\label{sec:axes}

With 84\% of the prefix's trainable parameters in the input encoder, parameter counts alone do not identify the mechanism; two modifications test it, and neither closes the gap.

\textbf{Initialization.} A zero-initialized gated residual may simply be easier to optimize than eight soft tokens competing inside the input softmax, so we gave the prefix the same branch ($\alpha_0 = 1.5$, $\Wout = 0$): it reads $0.7275 \pm 0.0221$ in Table~\ref{tab:arch}, slightly below the plain prefix and 6.6 points below \xattn{9}. The branch itself carries a trap that a hybrid hides. ReZero initializes residual gates to zero \citep{bachlechner2021rezero}, while Fixup initializes the last layer of each residual branch to zero \citep{zhang2019fixup}, but starting both the gate ($\alpha_0 = 0$) and the output projection ($\Wout = 0$) at zero closes every gradient path, so $(\alpha, \Wout) = (0, 0)$ is a fixed point of gradient-based training, AdamW's weight decay included; either choice alone is harmless, so the failure appears as an interaction. On Qwen2.5-1.5B the doubly-zero cell scores 0.277, 0.607 and 0.557 across seeds, with adapter gradient and gate exactly zero in the recorded diagnostic, yet the hybrid's prefix path keeps training, so a hybrid with a dead branch still improves on the benchmark. Appendix~\ref{app:init} gives the derivation, proved for function-preserving growth by \citet{khemais2026surgery}, and the ablation; Appendix~\ref{app:kv} describes a second trap in the deep key-value prefix.

\textbf{Conditioning length.} Prefix length is the axis the expressivity bound of \citet{petrov2024prompting} cares about, and positions come almost free: 8 to 64 slots costs 43,008 parameters, 0.03\% of the trainable total. Accuracy at seed 42 reads 0.7445, 0.7165 and 0.7439 at 8, 32 and 64 slots, a non-monotone spread smaller than the gap to the injected model.

An earlier pure-cross-attention pilot at 0.5B, with the input prefix removed, scores 0.6530, 0.6611 and 0.6704 at seeds 7, 42 and 123. It uses 4,000 training examples, 600 steps, learning rate $10^{-4}$ and 1,605 evaluation examples. Its scale and schedule differ from the main comparison, so it cannot isolate the prefix contribution at 3B.

\section{Hyperparameters}\label{app:hyper}

Table~\ref{tab:hyper} lists every setting needed to reproduce the runs. Values not listed are library defaults. A six-point learning-rate sweep from $10^{-3}$ to $10^{-5}$ (unaudited harness, 400 items, used only to choose the rate) places the optimum for both routes at an interior $3\times10^{-5}$, where the prefix reads 0.7300 against 0.7025 at $10^{-4}$ and 0.6850 at $10^{-5}$; both routes were then re-run at both candidate rates on the audited harness. Tuning the rate widens rather than closes the prefix--cross-attention gap: at seed 42 the prefix reads 0.7221 at $10^{-4}$ and 0.7445 at $3\times10^{-5}$, \xattn{9} reads 0.7601 and 0.7882, and the gap moves from 3.8 to 4.4 points (macro-F1: 7.6 to 9.7) in single runs, hence below the three-seed 5.5 of \S\ref{sec:sweep}.

\begin{table}[h]
\centering\footnotesize
\caption{Complete hyperparameter listing.}
\label{tab:hyper}
\begin{tabular}{@{}p{3.6cm}p{9.6cm}@{}}
\toprule
Setting & Value \\
\midrule
History encoder & \texttt{BAAI/bge-base-en-v1.5}, frozen, max 512 tokens per item \\
Control encoder & \texttt{BAAI/bge-base-en-v1.5}; robustness checks repeat the similarity assignment and the composite with \texttt{intfloat/e5-base-v2} \\
Input encoder & separate BGE encoder; all parameters trainable \\
Q-Former & 8 queries, 2 layers, width 768 unless varied in the sweep \\
Projection width & 1024 hidden units \\
Optimizer & AdamW, weight decay $10^{-3}$, gradient clipping at 1.0 \\
Schedule & linear with 0.05 warmup ratio \\
Precision & fp16 with loss scaling; SDPA attention; gradient checkpointing on \\
Model seeds & 7, 42, 123 \\
Control seed & 20260723, independent of the model seed \\
\midrule
\multicolumn{2}{@{}l}{\emph{Control battery, 3B (\S\ref{sec:control}, \S\ref{sec:peruser})}} \\
Decoder & Qwen2.5-3B-Instruct, frozen; \xattn{9} \\
Data & 8,000 train examples, 6,000 optimizer steps, batch size 1 \\
Evaluation & all 1,605 validation items, 323 users \\
Learning rate & $3\times10^{-5}$, the optimum selected in \S\ref{sec:protocol} \\
Controls & every history evaluated on each matched checkpoint \\
\midrule
\multicolumn{2}{@{}l}{\emph{Control battery, 0.5B (\S\ref{sec:robust}, \S\ref{sec:peruser})}} \\
Decoder & Qwen2.5-0.5B-Instruct, frozen; gated cross-attention \\
Data & 8,000 train examples, 6,000 optimizer steps, batch size 1 \\
Evaluation & all 1,605 validation items, 323 users \\
Learning rate & $3\times10^{-5}$; shared recipe, not a separate 0.5B tuning sweep \\
Shorter-schedule comparison & 4,000 examples, 600 steps, $10^{-4}$ (the schedule confound of \S\ref{sec:robust}) \\
\midrule
\multicolumn{2}{@{}l}{\emph{Audited model comparison (\S\ref{sec:axes})}} \\
Decoder & Qwen2.5-3B-Instruct, frozen; the recipe of the 3B control battery \\
Models & soft prefix at 8, 32 and 64 Q-Former queries; \xattn{9} \\
Seeds & 7, 42, 123 for the 8-slot prefix and \xattn{9}; 42 for the others \\
Serialization control & identical run with the constant \texttt{MOVIE:} label stripped from every profile item, seed 42 \\
\midrule
\multicolumn{2}{@{}l}{\emph{Initialization ablation (\S\ref{sec:axes}, Appendix~\ref{app:init})}} \\
Decoder & Qwen2.5-1.5B, frozen \\
Data & 4,000 train examples, 600 optimizer steps, batch size 1, 300 evaluation items \\
Learning rate & $10^{-4}$ \\
\midrule
\multicolumn{2}{@{}l}{\emph{Injection-site and tuning experiments (\S\ref{sec:sweep}, \S\ref{sec:prefix})}} \\
Decoder & Qwen2.5-3B, frozen \\
Data & 8,000 train examples, 6,000 optimizer steps, batch size 1 \\
Evaluation & 1,605 items for audited comparisons; 400 for the learning-rate grid \\
Learning rate & $3\times10^{-5}$ unless the sweep varies it \\
Injection layers & \texttt{range(18, 36, stride)} over the 36-layer decoder; strides 2, 4, 8, 18 give 9, 5, 3, 1 layers \\
\bottomrule
\end{tabular}
\end{table}

The other-task battery uses matched-trained 3B cross-attention checkpoints: 8,000 training rows, 6,000 steps, batch size 1, learning rate $3\times10^{-5}$. LaMP-1 and LaMP-4 use seeds 7 and 42; LaMP-3 and LaMP-5 use 7, 42 and 123. Evaluation uses 1,500 items except LaMP-3, which uses 1,605. Masked and random histories are evaluated on the corresponding matched checkpoint.

\section{Initialization trap: derivation and ablation}\label{app:init}

\textbf{Proposition.} Write the adapter branch of Eq.~\eqref{eq:xattn} as $A(h, Z) = \Wout\, C(h, Z)$ with $C(h, Z) = \mathrm{CrossAttn}(\mathrm{LN}(h), Z)$, so that $h' = h + \tanh(\alpha) A(h, Z)$, and let $g = \partial \mathcal{L} / \partial h'$. For any parameter $\theta$ of $C$ or of $\Wout$,
\begin{equation}
\frac{\partial \mathcal{L}}{\partial \theta} = \tanh(\alpha)\Big\langle g, \frac{\partial A}{\partial \theta}\Big\rangle,
\qquad
\frac{\partial \mathcal{L}}{\partial \alpha} = \big(1 - \tanh^2(\alpha)\big)\big\langle g, A(h, Z)\big\rangle.
\label{eq:grad}
\end{equation}
At $\alpha = 0$ the first expression is zero for every $\theta$; at $\Wout = 0$ we have $A \equiv 0$, so the second is zero as well. Under any optimizer whose update depends on the gradient alone, both parameters therefore keep their initial values at every step, and $(\alpha, \Wout) = (0, 0)$ is a fixed point the branch cannot leave. Decoupled weight decay does not lift it: it leaves $\Wout = 0$ at zero and $\alpha = 0$ at zero, so AdamW, which we use, leaves the branch dead. Neither condition alone is degenerate. With $\alpha = 0$ and $\Wout$ random, $A \neq 0$ makes $\partial \mathcal{L}/\partial \alpha$ generically non-zero, the gate moves off zero and the adapter gradients unlock; with $\Wout = 0$ and $\alpha_0 = 1.5$, $\tanh(\alpha) \neq 0$ keeps the adapter gradients open and $\Wout$ leaves zero on the first step. Only the conjunction deadlocks, which is why the failure appears as an interaction rather than a main effect of either initialization, and a parallel soft-prefix path keeps training throughout, which is how a hybrid model improves on the benchmark with a dead branch. The algebra is not new; \citet{khemais2026surgery} prove the same configuration for function-preserving network growth.
We contribute the measurement of what the failure costs and how it is detected.

\begin{table}[h]
\centering\small
\caption{Initialization ablation on LaMP-2 with Qwen2.5-1.5B, 600 steps, 300 evaluation examples, three seeds per cell, sample standard deviations. The doubly-zero branch has exactly zero gradient; the short-run accuracy differences are not a stable effect-size estimate. Gradient norm and gate magnitude are measured at step 590 of the seed-42 run.}
\label{tab:init}
\begin{tabular}{@{}lcc@{}}
\toprule
 & $\Wout = 0$ & $\Wout$ random \\
\midrule
Accuracy, $\alpha_0 = 0$ & $0.480 \pm 0.178$ & $0.677 \pm 0.033$ \\
Accuracy, $\alpha_0 = 1.5$ & $\mathbf{0.726 \pm 0.012}$ & $0.711 \pm 0.031$ \\
\midrule
Adapter gradient norm, $\alpha_0 = 0$ / $1.5$ & 0.0 / 5.7 & 2.1 / 7.4 \\
Gate $|\tanh \alpha|$, $\alpha_0 = 0$ / $1.5$ & 0.000000 / 0.904092 & 0.006787 / 0.903341 \\
\bottomrule
\end{tabular}
\end{table}

The doubly-zero cell scores 0.277, 0.607 and 0.557 at seeds 7, 42 and 123, respectively. At step 590 of seed 42, its adapter gradient norm and gate magnitude are exactly zero. The large spread in accuracy comes from a model whose parallel prefix still trains; a single run can therefore conceal the dead branch. This short ablation establishes the failure mode, not a reliable accuracy gain from changing initialization.

We use the gate magnitude deliberately narrowly. Gate magnitudes do not measure how much a branch contributes, because pre-gate activation scales vary with depth, and \citet{alayrac2022flamingo} say as much. Exact zero against non-zero asks something else, namely whether the parameter moved at all, which is a claim about optimization rather than influence.

\section{What an unaudited harness does to one model}\label{app:unaudited}

A harness that records no output validity changes the comparison asymmetrically. On the identical subset and seed it reproduces cross-attention to the digit, 0.8375, while placing the soft prefix at 0.7300 against the audited 0.7750 and its macro-F1 at 0.212 against 0.498. A model that looks as if it had learned the label prior and little else is being mis-parsed, and the gap between the two shrinks from 10.75 to 6.25 points once the parser is audited. This is why every prefix row in the main text was retrained under the audited harness.

\begin{figure}[h]
\centering
\includegraphics[width=0.48\textwidth]{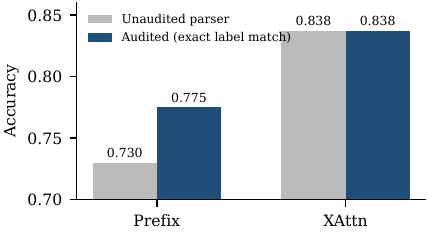}
\caption{Accuracy before and after auditing the parser, identical subset and seed: prefix 0.7300 to 0.7750, cross-attention 0.8375 in both cases. This diagnostic subset is separate from the full-validation architecture comparison.}
\label{fig:parser}
\end{figure}

\section{Injection routes and the deep key-value prefix}\label{app:kv}

\begin{figure}[h]
\centering
\resizebox{0.95\textwidth}{!}{%
\begin{tikzpicture}[x=1cm,y=1cm,font=\scriptsize,>={Stealth[length=1.6mm]},
  box/.style={draw,rounded corners=1pt,align=center,inner sep=2pt,line width=0.4pt},
  dec/.style={box,fill=black!6,minimum width=2.7cm,minimum height=1.7cm},
  mod/.style={box,fill=blue!12,minimum width=2.1cm,minimum height=0.4cm},
  xa/.style={box,fill=teal!18,minimum width=2.1cm,minimum height=0.4cm},
  kv/.style={box,fill=orange!22,minimum width=2.1cm,minimum height=0.32cm},
  zz/.style={box,fill=teal!18,minimum width=0.5cm,minimum height=0.5cm},
  qq/.style={box,fill=black!6,minimum width=2.4cm,minimum height=0.4cm}]
\begin{scope}
  \node[dec] (d) at (0,0) {};
  \node[anchor=north,font=\scriptsize] at (d.north) {Frozen decoder};
  \node[mod] (p) at (0,-0.4) {Soft prefix};
  \node[zz] (z) at (-2.0,0.1) {$Z$};
  \node[qq] (q) at (0,-1.4) {Query embeddings};
  \draw[->] (z) -- (p.west);
  \draw[->] (q) -- (d.south);
  \node[font=\scriptsize\bfseries] at (0,1.1) {(a) Prefix};
\end{scope}
\begin{scope}[xshift=4.5cm]
  \node[dec] (d) at (0,0) {};
  \node[anchor=north,font=\scriptsize] at (d.north) {Frozen decoder};
  \node[xa] (x) at (0,0.2) {Gated cross-attn ($\times k$)};
  \node[mod] (p) at (0,-0.4) {Soft prefix};
  \node[zz] (z) at (-2.0,0.1) {$Z$};
  \node[qq] (q) at (0,-1.4) {Query embeddings};
  \draw[->] (z) -- (x.west);
  \draw[->] (z) -- (p.west);
  \draw[->] (q) -- (d.south);
  \node[font=\scriptsize\bfseries] at (0,1.1) {(b) \xattn{$k$}: prefix retained + $k$ gated layers};
\end{scope}
\begin{scope}[xshift=9cm]
  \node[dec] (d) at (0,0) {};
  \node[anchor=north,font=\scriptsize] at (d.north) {Frozen decoder};
  \node[kv] (k1) at (0,0.3) {K/V prefix};
  \node[kv] (k2) at (0,-0.08) {K/V prefix};
  \node[kv] (k3) at (0,-0.46) {K/V prefix};
  \node[zz] (z) at (-2.0,0.1) {$Z$};
  \node[qq] (q) at (0,-1.4) {Query embeddings};
  \draw[->,dashed] (z) -- (k1.west);
  \draw[->,dashed] (z) -- (k2.west);
  \draw[->,dashed] (z) -- (k3.west);
  \draw[->] (q) -- (d.south);
  \node[font=\scriptsize\bfseries] at (0,1.1) {(c) Deep KV prefix};
\end{scope}
\end{tikzpicture}}
\caption{Memory injection routes; all three share the encoders and the Q-Former that produce $Z \in \mathbb{R}^{8\times768}$ from the history items and the current query. (a)~Prefix: $Z$ is projected and prepended to the input embeddings. (b)~\xattn{$k$}: the prefix is retained and $k$ decoder layers receive the gated update of Eq.~\eqref{eq:xattn}; this hybrid is the cross-attention model throughout. (c)~Deep key-value prefix, an appendix experiment. Decoder and history-item encoder are frozen; input encoder, Q-Former and injection modules train.}
\label{fig:routes}
\end{figure}
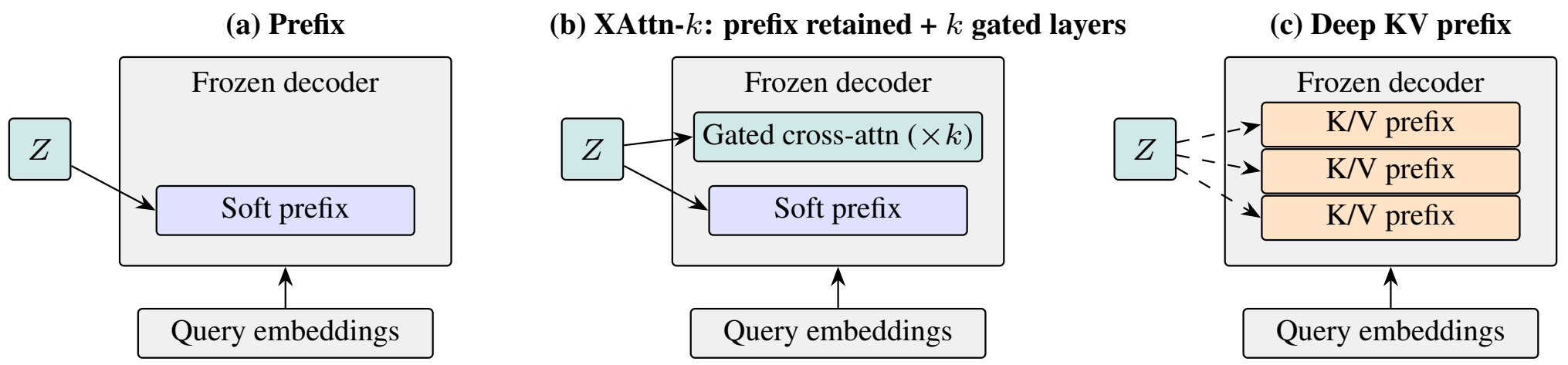

Figure~\ref{fig:routes} draws the two routes of the main text next to the third. The deep key-value prefix writes $Z$ into key-value prefixes at several decoder layers. Under this implementation it carries a second initialization trap, unrelated to the gate deadlock of Appendix~\ref{app:init}: its final projection starts at zero, so all $m$ slots begin identical and receive identical updates, and after 600 steps the maximum difference between slots is exactly zero, so lengthening the prefix adds copies of one state. Attention mass on the prefix rises from 0.191 at $m = 2$ to 0.296 at $m = 32$ while accuracy stays at 0.641, 0.641 and 0.645. With $m$ identical slots the softmax mass on the prefix grows with $m$ by construction, so the rise is expected and says nothing about use. Rising attention mass is not evidence of dilution here, since attention weights alone can misrepresent importance \citep{jain2019attention,serrano2019attention}. \citet{wiegreffe2019attention} argue that explanatory value depends on the definition and on diagnostic tests; \citet{kobayashi2020attention} show why transformed-value norms must also be considered.

\section{History as text, and the 0.5B decoder}\label{app:text}

Supplying the same four histories as text to the frozen decoder, with no training and no compression, gives 0.6617 Matched, 0.6386 Similar-user, 0.6069 History masked and 0.5832 Random-user on LaMP-2. Absolute numbers are not comparable with the adapter, which is trained on the task, but the ordering and the relative sizes are (Appendix~\ref{app:text}). The same battery on a frozen Qwen2.5-0.5B trained to the recipe of Appendix~\ref{app:hyper} gives 0.7751 Matched, 0.7304 Similar-user, 0.7232 History masked and 0.6467 Random-user: Matched minus Similar-user is 4.5 points $[2.6, 6.6]$, Similar-user minus History masked 0.7 points $[-1.1, 2.7]$, and History masked minus Random-user 7.6 points. A 0.5B checkpoint trained on the shorter schedule (4,000 examples, 600 steps, learning rate $10^{-4}$) shows almost no penalty under any replacement; the large discrepancy shrinks when the recipe is matched, although paired permutation tests still distinguish the scales (\S\ref{sec:peruser}). On the other tasks the same text histories hurt: LaMP-3 mean absolute error is 0.745 with no history against 1.032 with the matched history and 1.574 with a random one, mostly through format, since the numeric-output rate falls from 0.973 to 0.90 while the error on in-range outputs alone stays between 0.666 and 0.719, and LaMP-5 ROUGE-1 falls from 0.353 without a history to 0.188 with the matched history.

\section{Training-loss estimates}\label{app:loss}

At batch size 1 a per-step loss is heavy-tailed, so the tail mean of a training log is a noise estimate rather than a fit measurement. The result files keep every tenth step; over the final third of training, 200 logged points per run, the mean and median across three seeds are: prefix 2-layer $0.489 \pm 0.079$ and 0.0041; 4-layer $0.485 \pm 0.075$ and 0.0039; 8-layer $0.582 \pm 0.054$ and 0.0126; 16-layer $0.724 \pm 0.176$ and 0.0283; gated prefix $0.529 \pm 0.074$ and 0.0087; \xattn{1} $0.366 \pm 0.024$ and 0.0011; \xattn{9} $0.330 \pm 0.111$ and 0.0012. The within-run standard error of the mean is near 0.08 for every model, which is the size of the between-model differences it would have to resolve, so Table~\ref{tab:arch} reports the median. Both statistics place the injected models below the prefix models in training loss.

\section{Expressivity against optimization, and gradient flow into the Q-Former}\label{app:grad}

\citet{li2021prefix} report that longer prefixes eventually reduce training loss but slightly worsen test performance, which is consistent with overfitting, and \citet{lester2021power} find diminishing gains and mild degradation at longer prompt lengths. The decline in \S\ref{sec:prefix} has a different signature, a training loss that rises with capacity, and two mechanisms could produce it. One is an expressivity ceiling. Within one attention head with fixed content representations, a prefix preserves relative content attention and adds a bias in a subspace of dimension at most its length \citep{petrov2024prompting}; this is not a decoder-wide bound, since later-layer attention can change, and \S\ref{sec:axes} shows that eight times the conditioning positions does not move the prefix. The other is plain optimization failure, of which a rising training loss is the usual signature. The gradient diagnostics below are consistent with both.

The models use the same Q-Former design, so we can ask how much learning signal each injection route delivers to it. During training we log the gradient norm reaching the Q-Former and the gradient norm reaching the injection module, and divide each by the square root of that module's parameter count so that the comparison does not merely reflect module size (Table~\ref{tab:grad}).

\begin{table}[h]
\centering\small
\caption{Root-mean-square gradient per parameter reaching the Q-Former and the injection module, averaged over the final quarter of training, seed 42.}
\label{tab:grad}
\begin{tabular}{@{}lrrcc@{}}
\toprule
Model & Q-Former params & Injection params & Q-Former & Injection \\
\midrule
Prefix, 2-layer & 18,909,696 & 1,579,008 & $1.82 \times 10^{-3}$ & $9.28 \times 10^{-3}$ \\
Prefix, 16-layer & 118,139,904 & 1,579,008 & $5.36 \times 10^{-4}$ & $1.35 \times 10^{-2}$ \\
\xattn{1} & 18,909,696 & 19,947,521 & $1.03 \times 10^{-3}$ & $1.75 \times 10^{-3}$ \\
\bottomrule
\end{tabular}
\end{table}

The Q-Former-to-injection ratios are 0.196, 0.040 and 0.589. The cross-attention ratio is about fifteen times the deep prefix's ratio, whereas its Q-Former gradient itself is only about $1.9\times$ as large ($1.03 \times 10^{-3}$ against $5.36 \times 10^{-4}$). Deepening the prefix reduces the Q-Former value by about $3.4\times$ and increases the injection-module value. These diagnostics accompany the three-seed macro-F1 decline from 0.490 to 0.375 in Table~\ref{tab:arch}; they do not establish that one module causally absorbs another's learning signal. We report this as one measurement at one seed on one task. It is consistent with the expressivity account of \citet{petrov2024prompting} and equally with optimization failure, and separates neither.

\section{Position relative to prior and concurrent controls}\label{app:position}

Table~\ref{tab:position} sets this paper's control battery beside the replacement controls used by the closest prior and concurrent work. The cells summarize each paper's own description; a dash means the item is not reported there. PRIME swaps user-specific LoRA weights as well as text, so its effective checkpoint changes. The comparison column includes non-personalized baselines when they differ from an input mask.

\begin{table}[h]
\centering\footnotesize
\setlength{\tabcolsep}{3pt}
\caption{Replacement controls in prior and concurrent work and in this paper.}
\label{tab:position}
\begin{tabular}{@{}>{\raggedright\arraybackslash}p{2.0cm}>{\raggedright\arraybackslash}p{2.3cm}>{\raggedright\arraybackslash}p{2.5cm}>{\raggedright\arraybackslash}p{1.5cm}>{\raggedright\arraybackslash}p{1.4cm}>{\raggedright\arraybackslash}p{1.4cm}>{\raggedright\arraybackslash}p{1.3cm}@{}}
\toprule
Work & Setting & Replacement & Graded by similarity & Fixed checkpoint & Contrast vs.\ no or masked history & Negative rebuilt \\
\midrule
LaMP \citep{salemi2024lamp} & retrieval of text items & random items from the user's own profile, or from the whole corpus & two levels & zero-shot and fine-tuned & yes & -- \\
\citet{wu2024profile} & retrieval of text items (Flan-T5 fine-tuned; frozen Llama-2) & query-relevant items from a pool of other users' items & by query relevance, not user similarity & trained per augmentation & yes & -- \\
PRIME \citep{zhang2025prime} & textual profile plus per-user LoRA & whole profile and LoRA of another user & four levels (most similar to dissimilar) & no: LoRA swapped & Non-P & -- \\
ENOUGH \citep{liu2026enough} & retrieval of text records, frozen generator & one record replaced by a content-matched other-user record; also random user and whole-profile swap & matched on context, relevance, length, time & yes & yes & matching audit \\
\citet{zheng2026weighting} & multimodal recommenders & shuffle (permutation) of per-user weights, plus a population-weight contrast and a signal-implant dose--response & dose--response & yes & -- & -- \\
This paper & eight-vector compressed memory around a frozen decoder, and raw text & whole profile: random user, similarity-matched user, composite of other users' items & similarity and dose curves; category matching & yes & yes, masked and trained & two encoders, three greedy builds \\
\bottomrule
\end{tabular}
\end{table}

\section{Per-seed and per-construction contrasts}\label{app:perseed}

Table~\ref{tab:perseed} unpacks the means of Table~\ref{tab:battery} by seed and adds the alternative constructions of the negative, so that the spread behind every headline number is visible. Entries are differences of the per-seed accuracies in Table~\ref{tab:battery}, in points; the bootstrap point estimates quoted in the main text can differ from the mean of the three seeds in the last digit.

\begin{table}[h]
\centering\small
\caption{Contrasts by seed (points) on the 3B \xattn{9} checkpoints, and descriptive ratios relative to masking (not the trained no-history comparator). The last block gives the alternative constructions of the negative: two greedy rebuilds of the composite at seed 42, and the Similar-user and composite negatives rebuilt under the E5 encoder (three-seed means).}
\label{tab:perseed}
\begin{tabular}{@{}lcccc@{}}
\toprule
Contrast & Seed 7 & Seed 42 & Seed 123 & Mean \\
\midrule
Matched $-$ History masked (gain over masked history) & 5.5 & 4.4 & 4.5 & 4.8 \\
History masked $-$ Random-user (harm) & 5.2 & 4.4 & 6.2 & 5.3 \\
Matched $-$ Random-user & 10.7 & 8.8 & 10.8 & 10.1 \\
Matched $-$ Similar-user & 4.2 & 2.4 & 4.1 & 3.6 \\
Matched $-$ Composite ($\cos = 0.90$) & 3.8 & 2.9 & 3.9 & 3.5 \\
Matched $-$ Composite ($\cos = 0.95$) & 1.2 & 0.8 & 1.5 & 1.2 \\
Similar-user $-$ History masked & 1.2 & 2.1 & 0.4 & 1.2 \\
\midrule
Harm share & 0.49 & 0.50 & 0.58 & 0.52 \\
Residual share, Similar-user & 0.77 & 0.53 & 0.90 & 0.74 \\
Residual share, Composite ($\cos = 0.95$) & 0.22 & 0.18 & 0.33 & 0.24 \\
\midrule
\multicolumn{5}{@{}l}{\emph{Alternative constructions of the negative}} \\
Matched $-$ Composite, greedy rebuilds at seed 42 & \multicolumn{4}{l}{0.8 (original), 0.6, 1.2; mean 0.9, SD 0.3} \\
Matched $-$ Similar-user under E5 & \multicolumn{4}{l}{4.0 (three-seed mean)} \\
Matched $-$ Composite under E5 (mean cosine 0.978) & \multicolumn{4}{l}{1.7 (three-seed mean)} \\
\bottomrule
\end{tabular}
\end{table}

The seed-level values qualify the averages. Here harm share is (masked $-$ random)/(matched $-$ random), and residual share is (matched $-$ negative)/(matched $-$ masked). The residual share for the Similar-user control ranges from 0.53 to 0.90 across seeds, so its mean is sensitive to the seed and to the masked reference; and seed 42, the seed used for the single-run comparisons elsewhere in the paper, is the one on which every contrast is smallest.

\begin{table}[h]
\centering\small
\caption{Matched minus each control, accuracy points. The $t$-interval uses three seed-level contrasts; the bootstrap resamples seeds and users (10,000 replicates).}
\label{tab:intervals}
\begin{tabular}{@{}lrrrrr@{}}
\toprule
Control & Seed 7 & Seed 42 & Seed 123 & $t$, 2 df & Bootstrap \\
\midrule
Random-user & 10.72 & 8.85 & 10.78 & $[7.39,12.84]$ & $[8.01,12.30]$ \\
Similar-user & 4.24 & 2.37 & 4.11 & $[0.98,6.17]$ & $[2.09,5.23]$ \\
Composite, 0.95 & 1.18 & 0.81 & 1.50 & $[0.31,2.02]$ & $[0.40,1.89]$ \\
Composite, labels & 1.31 & 0.37 & 1.06 & $[-0.29,2.12]$ & $[0.09,1.79]$ \\
History masked & 5.48 & 4.42 & 4.55 & $[3.38,6.26]$ & $[3.41,6.34]$ \\
Own labels only & 2.93 & 1.93 & 3.24 & $[1.00,4.40]$ & $[1.25,4.21]$ \\
\bottomrule
\end{tabular}
\end{table}

The composite construction grid (original, rebuild 2, rebuild 3) is $(1.18,0.81,1.50)$, $(1.87,0.62,1.81)$ and $(2.12,1.18,1.43)$ points in seed order 7, 42, 123. Resampling seeds, constructions within seeds, and users gives 1.39 points with interval $[0.74,2.05]$. The trained no-history comparison is between separately trained checkpoints: Matched minus this comparator is 2.06, 2.12 and 2.62 points.

\section{Algorithms}\label{app:algorithms}

The following procedures define the replacement histories and null distributions. Profile embeddings are BGE embeddings of the serialized items; a profile centroid is the mean of its item embeddings.

\paragraph{Similar-user derangement.} Form the $U \times U$ matrix of cosines between profile centroids, set the diagonal to $-\infty$, and solve the linear assignment problem that maximizes the total cosine (\texttt{scipy.optimize.linear\_sum\_assignment} on the negated matrix). The solution is a permutation with no fixed point; each user receives the history of the user it is assigned to. The same mapping is used at every seed.

\paragraph{Composite negative.} For a target profile with centroid $c$: (1) rank all items owned by other users by cosine to $c$ and keep the 400 nearest as the candidate pool; (2) starting from the empty set, add six times the candidate that brings the centroid of the selected items closest to $c$; (3) serialize the six items as a history. The stochastic variant replaces the argmax in step (2) by a uniform draw among the five best candidates, seeded by the run seed plus the user's index, which is how the rebuilds of Table~\ref{tab:perseed} were produced. For the cosine-0.90 composite, all other-user items are candidates and each greedy step minimizes the absolute difference between the running centroid cosine and 0.90.

\paragraph{Label-matched composite.} Recover the category of each of the first six raw profile items. For each position, exclude the target user's items and any item whose article text equals an evaluation query. Among items of the required category, retain the 400 nearest to the target centroid and greedily select the unused item maximizing the running centroid cosine. Retain the donor title and its own category; no query answer is used. Record fallbacks if a category has no candidate (none occurred).

\paragraph{Labels-only and dose controls.} Serialize each category as \texttt{MOVIE: "<category>"}, using either the matched profile's labels or the random donor's. For dose replacement, seed a permutation of six positions by $20261017$ plus the sorted user index, and replace its first $k$ positions with the corresponding positions from a fixed random, similar or label-matched donor. The subsets are nested and shared across model seeds. The endpoints reuse the matched and full-replacement evaluations.

\paragraph{Zero-effect null.} We use 2,000 replicates for each null. For each validation item, exchange the outcomes recorded under the two conditions with probability one half, independently across items; recompute $\Delta_u$ for every user and then the Gini of $|\Delta_u|$ and the standard deviation of $\Delta_u$; repeat. The permutation probability is the fraction of replicates whose statistic is at least the observed one.

\paragraph{Homogeneous-effect null.} Pool the item-level outcome pairs of all users; for each user draw with replacement as many pairs as the user has items; recompute the same statistics; repeat. This keeps the population effect and the item counts and removes only the assignment of items to users.

\section{Applying the audit to another system}\label{app:recipe}

The fixed-checkpoint battery requires an evaluation split and an embedding model; the trained no-history and user-disjoint comparisons additionally require training.
\begin{enumerate}
\item Audit train/evaluation user overlap and reserve users for a disjoint evaluation. Train a no-history comparator under the same recipe.
\item Check branch gradients at initialization and report exact-label parsing and valid-output rates.
\item Evaluate fixed checkpoints with matched, random, similar-user, composite and masked histories. When labels exist, add labels-only and label-matched controls; never use the query's answer to construct a donor.
\item Report matched-minus-control contrasts separately from the gain over the trained no-history model. Neither a swap penalty nor masking alone measures deployment benefit.
\item Rebuild negatives across encoders and construction seeds. Include construction variability in uncertainty, and cluster similarity curves by user.
\item Replace nested subsets of history items to measure within-user responses. Check query/profile duplicates and evaluate per-user concentration against nulls at the observed query counts.
\end{enumerate}

\section{Completed controls and remaining tests}\label{app:threats}

The revision adds positional category matching, labels-only histories, article-level near-duplicate filtering, nested within-user replacement, a complete seed-by-construction grid, user-disjoint training, and a movie-tagging pilot. Sections~\ref{sec:labels}--\ref{sec:unseen} report their results. These checks leave uncertainty about small gains on new users and about the representation learned from a stable donor profile. More seeds, longer histories, and a trained no-history comparator within each user fold would strengthen the generalization test. For the architecture comparison, a pure cross-attention arm at 3B, projection-width sweeps and separate prefix optimization remain untested.


\end{document}